\pdfoutput=1
\documentclass{article}
\usepackage{iclr2026_conference,times}
\usepackage{amsmath,amsfonts,bm}

\def\figref#1{figure~\ref{#1}}

\def\secref#1{section~\ref{#1}}

\def\twosecrefs#1#2{sections \ref{#1} and \ref{#2}}

\def\eqref#1{equation~\ref{#1}}

\def\1{\bm{1}}

\DeclareMathAlphabet{\mathsfit}{\encodingdefault}{\sfdefault}{m}{sl}
\SetMathAlphabet{\mathsfit}{bold}{\encodingdefault}{\sfdefault}{bx}{n}

\usepackage[hidelinks]{hyperref}
\usepackage{url}
\usepackage{graphicx}
\usepackage{booktabs}
\usepackage{tabularx}
\newcolumntype{Y}{>{\raggedright\arraybackslash}X}
\usepackage{xcolor}
\usepackage{microtype}
\usepackage{wrapfig}
\usepackage{capt-of}
\usepackage{enumitem}
\setlist{leftmargin=1.3em,topsep=2pt,itemsep=1pt,parsep=0pt}

\newcommand{\pkgname}{mjfit}
\newcommand{\mi}{\texttt{\pkgname}}
\newcommand{\modebox}[1]{\textcolor[HTML]{#1}{\rule{0.6em}{0.6em}}\,}
\providecommand{\tabref}[1]{Table~\ref{#1}}

\title{Blind Precise Assembly: Proprioceptive\\Sim-to-Real Peg-in-Hole for Diverse Parts}
\author{Antonia Bronars\thanks{Equal contribution.} \quad Younghyo Park\footnotemark[1] \quad Kushal Kedia
\quad Pulkit Agrawal \\
\normalfont Massachusetts Institute of Technology}
\iclrfinalcopy

\begin{document}
\maketitle
\lhead{Preprint}

\begin{abstract}
Every robot already carries high-resolution contact sensors: its own joints.
Legged robots and humanoids exploit them to walk, run, and track whole-body
motions without vision, but there the contacts are large and forceful. Is joint sensing also enough for precise, contact-rich assembly, where the robot must
find a millimeter-scale opening, mesh gear teeth, or catch a thread through light
contact? We study this question by training blind assembly policies that observe
only a short history of joint positions, velocities, and their own actions, without
visual, force, torque, or tactile sensing. The poses of both the fixture and the part
are unknown to the policy, and the grasp is randomized substantially. The policy must
therefore assemble the parts through contact, sensed exclusively through its own
proprioception. Trained with reinforcement learning entirely in simulation over
randomized parts, grasps, and contacts, such policies learn peg insertion, gear
meshing, and nut threading. The peg insertion policy transfers zero-shot to the real
world and inserts pegs of different shapes and sizes in 73\% of trials.
Controlled comparisons with matched visual policies in simulation show that blind policies generalize far better to unseen part shapes and sizes, though they can be slower.
Training such policies needs thousands of different assemblies, each
simulated with its clearance intact, so we also release an
open-source generator that builds them procedurally from exact convex pieces
for MuJoCo; it is more accurate and faster than convex decomposition or
MuJoCo's signed-distance collision. Videos and interactive simulations are at
\href{https://younghyopark.me/blind-proprioceptive-assembly/}{\nolinkurl{younghyopark.me/blind-proprioceptive-assembly/}}.
\end{abstract}

\section{Introduction}

A force applied anywhere on a robot's body travels through its links to the
joints, which then move differently from what their commands alone would
produce. The encoders record that difference, so the joints of every arm and
leg double as contact sensors. Legged robots exploit this proprioceptive
signal to walk, run, and climb stairs blind
\citep{Lee2020Quadrupedal,Siekmann2021Gaits,Siekmann2021BlindStairs,Kumar2021RMA},
and humanoids to walk and track whole-body motions
\citep{Radosavovic2024Humanoid,Cheng2024Expressive,He2024OmniH2O}. In these
settings, however, contacts are large and forceful, and joints often comply meaningfully while recording measurable deviations. Can robots exploit the same signal for precise, contact-rich assembly?

Precise assembly with a robotic arm occupies the opposite regime: the robot must
find an opening with a millimeter or less of clearance, bring gear teeth into mesh, or
catch a thread, through light contact whose effect on the joints is easy to miss. We
study this question by training \emph{blind} assembly policies that have no
information outside of their own joint sensing, and verify that proprioception
contains enough information for three such tasks: peg insertion, gear meshing, and
nut threading (\figref{fig:tasks}). In the real world, our peg insertion policy
inserts eight pegs of different cross-sections, three of them outside
its training distribution, with 0.5--1.5\,mm of clearance per side, in
73\% of 237 trials, without being
given the hole's location, the peg's pose in its grasp, the part's size, or
its clearance (\secref{sec:real}). To our knowledge,
this is the first such demonstration.

\begin{figure}[t]
  \centering
  \includegraphics[width=\linewidth]{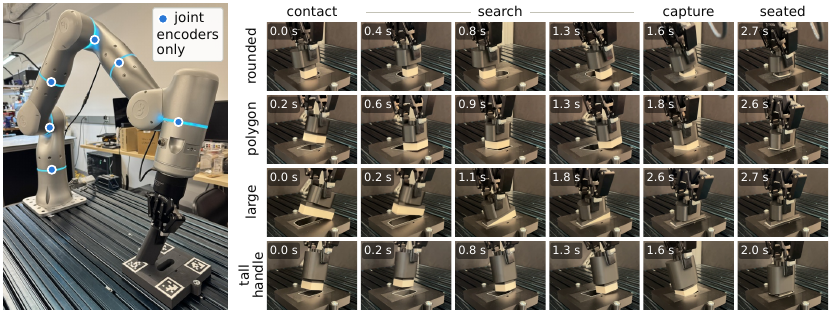}
  \caption{\textbf{Blind insertion.} A policy trained only in simulation
  inserts pegs of different shapes and sizes in the real world using only its
  joint encoders and previous actions.}
  \label{fig:teaser}
\end{figure}

We train our policies entirely in simulation with reinforcement learning and
deploy the peg insertion policy in the real world without fine-tuning
(\figref{fig:teaser}). Training
randomizes part geometry, contact properties, fixture poses, and grasps, which can
hold the part off-center on the finger pads and tilted in the grasp. The policy
observes a 0.6\,s history of joint positions and velocities, the gripper
pose and twist computed from them, and its own previous actions. It receives
no image and no force signal, so it can perceive contact and find the hole, shaft,
or bolt only through its own motion (\secref{sec:method}).

On all three tasks, probing the blind policies' learned representations and comparing them with matched visual policies revealed two interesting properties of visually blind behaviors:
\begin{itemize}
  \item \textbf{Visually blind yet intelligent.} The learned behavior is not an open-loop pattern: its learned representation encoding its own proprioceptive inputs contains meaningful task-relevant information that the policy itself heavily relies on to adjust its behavior (\secref{sec:smart}).
  \item \textbf{Proprioception generalizes better.} Because a blind policy depends on the part only through its contacts, it degrades far less than vision on part shapes and sizes it has not seen during training (\secref{sec:unseen}).
\end{itemize}

Training for this robustness requires diverse parts: thousands of different
assemblies, generated procedurally and simulated in parallel, each with its
millimeter clearance intact. In MuJoCo \citep{Todorov2012MuJoCo}, building
holes from convex primitives by hand does not scale to random shapes, and the
automatic alternatives, convex decomposition and signed-distance-field (SDF)
collision, are less accurate and slower at these clearances. We therefore
built \mi{}, which procedurally generates pegs and holes from any planar outline, as well as threads and gears, out of exact convex prisms. To summarize, our contributions are:

\begin{enumerate}[leftmargin=1.5em]
  \item Blind policies, trained purely in simulation, that learn peg insertion,
  gear meshing, and nut threading from proprioception alone, given no fixture pose,
  grasp, or part model
  (\secref{sec:existence}), with analyses showing that
  on all three tasks they search reactively,
  acting on the contact state their joint histories encode rather than following a
  fixed pattern (\secref{sec:smart}), and that they generalize to unseen part shapes and
  sizes better than matched visual policies (\secref{sec:unseen});
  \item The first sim-to-real policy for precise insertion from proprioception alone:
  trained only in simulation, the peg insertion policy transfers zero-shot to a real
  robot and inserts pegs of different cross-sections, including pegs outside its
  training distribution (\secref{sec:real});
  \item \mi{}, an open-source procedural generator of assemblies for MuJoCo
  (pegs and holes, threads, and gears) with exact convex collision geometry,
  which supplies the thousands of different training assemblies, and a
  benchmark against analytic ground truth in which it is more accurate and
  faster than convex decomposition and MuJoCo's SDF collision
  (\secref{sec:sim}).
\end{enumerate}

\section{Learning blind assembly}
\label{sec:method}

We train blind policies for three assemblies that \mi{} generates: peg insertion, gear
meshing, and nut threading. We describe the tasks (\secref{sec:task}); how \mi{}, our
open-source procedural generator for MuJoCo, supplies thousands of different
assemblies with exact convex collision geometry that preserves their clearances
(\secref{sec:sim}); and the policy, whose auxiliary contact supervision makes learning
faster without giving the actor any privileged information
(\secref{sec:policy}; \figref{fig:aux}).

\subsection{Tasks}
\label{sec:task}

\begin{figure}[t]
  \centering
  \includegraphics[width=\linewidth]{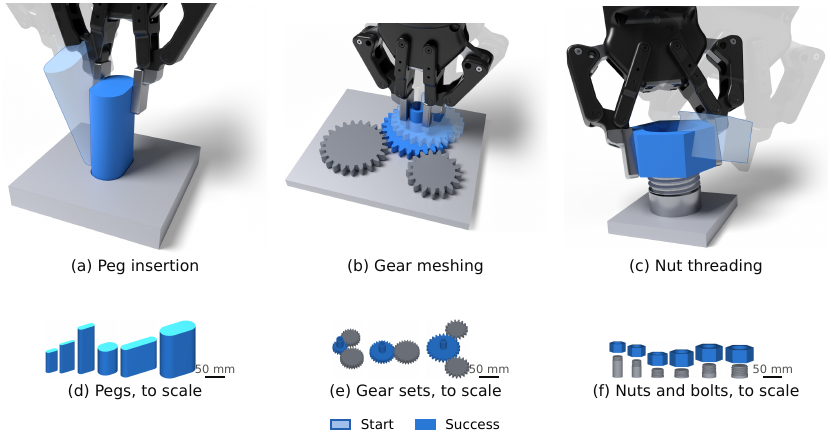}
  \vspace{-1.4em}
  \caption{\textbf{Peg insertion, gear meshing, and nut threading.} (a--c)~A start of
  each task, faded, and the same grasp at success: the peg seated in
  its hole, the gear seated on its shaft in mesh with both anchor gears, and the nut
  turned until its thread has caught. (d--f)~Stadium pegs, gear sets, and nuts and
  bolts generated by \mi{} and used for training, to scale.}
  \label{fig:tasks}
\end{figure}

\begin{wrapfigure}{r}{0.31\linewidth}
  \centering
  \vspace{-1.2em}
  \includegraphics[width=\linewidth]{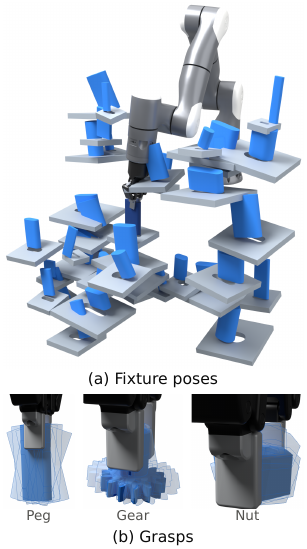}
  \vspace{-1.6em}
  \caption{\textbf{Training fixture poses and grasps.} Gear and bolt fixtures span
  the peg's ranges.}
  \label{fig:workspace}
  \vspace{-1.0em}
\end{wrapfigure}

In all three tasks (\figref{fig:tasks}), a 7-DOF robot arm equipped with a parallel jaw gripper holds a part
and must assemble it with a fixture that can sit almost anywhere in its workspace, and tilted (\figref{fig:workspace}a). The part's geometry, its
pose in the grasp (\figref{fig:workspace}b), and the contact properties are randomized and not observed by policy (\tabref{tab:ranges}). Episodes
start with the part touching the fixture but misaligned, and succeed once it is
fully assembled. This requires the policy to both search for alignment through contact, and then perform the assembly, based on its joint position feedback alone.
\begin{itemize}
    \item \emph{Peg insertion:} the pegs are prisms with varying cross-section, size,
aspect ratio, mass, and length, with 1.0--1.5\,mm of clearance per side
(Appendix~\ref{app:training}). Pegs start typically around 2\,cm from the hole,
tilted far more than the clearance admits, and succeed when the peg is seated in the
hole.
\item \emph{Gear meshing:} spur gears of varying size and tooth count must slide down a
shaft into mesh with one or two anchor gears that turn on their own axles, with
0.2--0.6\,mm of backlash. Episodes start with the gear in contact with the shaft tip, but misaligned and with its teeth out of phase with the anchors', and succeed when the gear
is seated on the shaft with teeth meshed.
\item \emph{Nut threading:} metric nuts from M24 to M42 must thread onto a bolt with 0.05--0.2\,mm of thread clearance. Episodes start with the nut resting on the bolt's
tip, off its axis and tilted, and succeed a quarter turn after the thread catches.
\end{itemize}

\subsection{Simulating tight insertion at scale}
\label{sec:sim}

Training on this distribution means simulating thousands of parallel worlds
that each hold a different, randomly generated assembly, with per-side
clearances down to 0.5\,mm. MuJoCo collides meshes
through their convex hulls, so a concave hole must be given an SDF or built
from convex pieces, which simulation benchmarks assemble from boxes by hand
\citep{Zhu2020robosuite,Tao2024ManiSkill3,Zakka2025Playground}.

\begin{figure}[t]
  \centering
  \includegraphics[width=\linewidth]{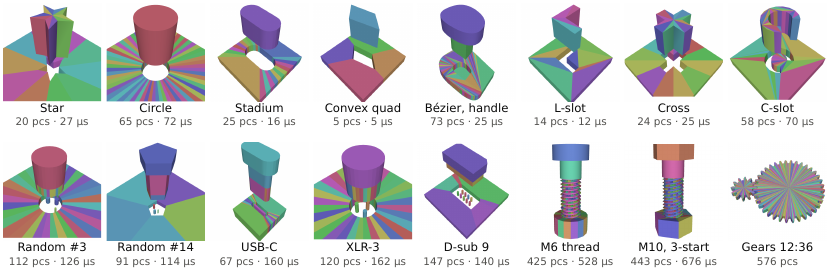}
  \vspace{-1.2em}
  \caption{\textbf{Assets generated by \mi{}} one color per convex
  piece of the generated collision geometry. Labels give the number of convex pieces (pcs) and the CPU time per
  physics step with the parts in contact.}
  \label{fig:assets}
\end{figure}

\paragraph{Exact convex decomposition.} \mi{} automates the hand-built hole
and generalizes it to arbitrary outlines. Given the planar outline of a hole,
optionally with islands such as connector pins, it emits a ready-to-simulate
MJCF model in which every contact surface is a MuJoCo primitive or a convex
prism (\figref{fig:assets}). The plate around the hole is split into convex
polygons and extruded; the peg is the hole outline offset inward by the
clearance, so every peg edge sits exactly the clearance from its hole edge,
and islands become pockets in a hollow peg. Curves are tessellated by a
\emph{circumscribed} polygon, so the wall never intrudes into the true hole,
as it does in rings of boxes that inscribe the circle
\citep{Zhu2020robosuite}; the only error is extra play at the corners,
$c\,(1/\cos(\pi/n)-1)$ for a circle of $n$ sides at radial clearance $c$.
Because the construction knows which region must stay empty, the user
specifies the clearance directly rather than tuning a concavity tolerance and
checking the result. Companion generators apply the same idea to bolt--nut
pairs and spur gears, and a sampler draws random outlines, islands, and
dimensions, so one training run can hold thousands of different assemblies.

\begin{wrapfigure}{r}{0.48\linewidth}
  \centering
  \vspace{-1.2em}
  \includegraphics[width=\linewidth]{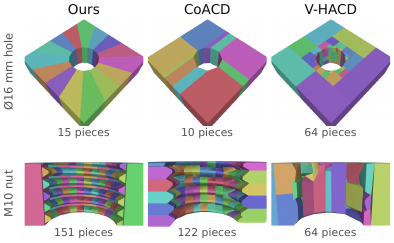}
  \vspace{-1.6em}
  \caption{\textbf{Same parts, three decompositions}, one color per convex
  piece (Ø16\,mm plate; M10 nut cut open). Ours errs only toward extra play;
  CoACD and V-HACD put material inside the hole and fill the nut's
  0.1\,mm thread clearance.}
  \label{fig:decomp}
  \vspace{-1.2em}
\end{wrapfigure}

\begin{figure}[t]
  \centering
  \includegraphics[width=\linewidth]{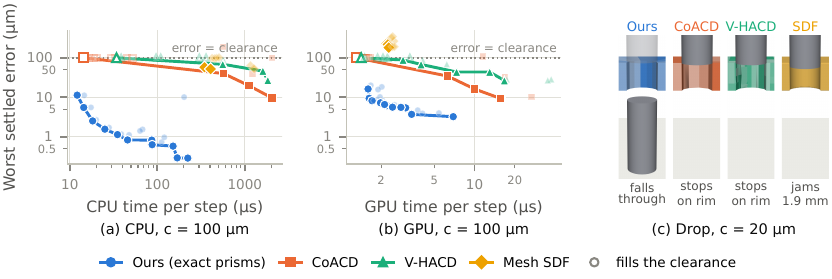}
  \vspace{-1.5em}
  \caption{\textbf{Speed against accuracy} for a 16\,mm round peg with
  100\,\textmu m clearance, (a) on CPU, one environment on one thread, and (b)
  on GPU, per environment step across 256 MuJoCo Warp worlds, with the error
  also measured on Warp. Each line is one method's Pareto front over all its
  settings; lower left is better, and hollow markers fill the clearance (dotted
  line).
  (c) The peg, centered and aligned, dropped from 1\,mm above each method's
  default plate and shown 0.2\,s later.}
  \label{fig:pareto}
\end{figure}

\paragraph{Benchmark.} We compare against the three automatic ways to give
MuJoCo a concave hole: CoACD \citep{Wei2022CoACD}, V-HACD
\citep{Mamou2009VHACD}, and MuJoCo's native mesh SDF, each swept over its
granularity setting. The test is a 16\,mm round or square hole with clearance
down to 20\,\textmu m; we measure
each method's error against its time per step, on CPU and on GPU with MuJoCo
Warp \citep{MuJoCoWarp} (\figref{fig:pareto}; Appendix~\ref{app:bench}).
Approximate decomposition tolerates concavity far larger than an assembly's
clearance, so it leaves wall material inside the hole (\figref{fig:decomp});
refining it narrows the error but never closes it (\figref{fig:pareto}a).
Mesh SDF misses contacts and rounds the rim. These errors change outcomes: a
centered peg stops on the rim or jams on every baseline's default plate, but
falls cleanly through ours (\figref{fig:pareto}c). \mi{} is also an order of
magnitude faster than mesh SDF on CPU at equal or better accuracy
(\tabref{tab:speed}), and has the best speed--accuracy trade-off on GPU
(\figref{fig:pareto}b). Each assembly is only a few dozen convex pieces, so
every one of our 12,288 parallel training worlds \citep{mjlab} can hold a
different one.

\subsection{Policy and training}
\label{sec:policy}

\begin{figure}[t]
  \centering
  \includegraphics{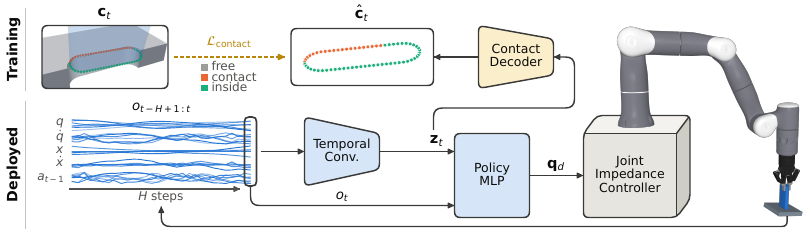}
  \vspace{-1.2em}
  \caption{\textbf{Policy Architecture}. A temporal convolution compresses the last $H=30$ proprioceptive observations $o_t$ (joint positions $q$ and velocities $\dot q$, gripper pose $x$
  and twist $\dot x$, and the previous action $a_{t-1}$) into a latent $z_t$,
  and an MLP maps $z_t$ and the current observation $o_t$ to joint
  targets $q_d$ for a joint impedance controller. During training, we provide contact supervision from privileged simulator state. A contact decoder (amber) predicts the contact state ($\hat c_t$) from $z_t$. Its cross-entropy $\mathcal{L}_\mathrm{contact}$ with the
  simulator's labels $c_t$ (left, near half of the fixture cut away) trains
  the encoder, but $\hat c_t$ never reaches the policy. }
  \label{fig:arch}
\end{figure}

\paragraph{Policy.} At each step, the actor (\figref{fig:arch}) observes its
joint positions and velocities, the gripper pose and twist computed from them
by forward kinematics, and its previous action. It never receives the hole
pose, the part geometry, a wrench estimate, or an image. A temporal
convolutional encoder compresses the last 0.6\,s of observations into a
latent, and an MLP maps the latent and the current observation to
joint-position increments. A joint impedance controller tracks them on an arm model
whose armature, damping, and friction were identified from real-world measurements.
Additional training details in Appendix~\ref{app:training}.

\paragraph{Training.} We use PPO \citep{Schulman2017PPO} with a sparse
reward: a bonus on success and one-time bonuses for intermediate milestones, such
as the peg's first capture by the opening (Appendix~\ref{app:gear-nut} gives the
gear's and nut's). Two training-only signals speed up learning from the sparse reward
without giving the actor privileged information: an asymmetric critic
\citep{Pinto2018Asymmetric} that also sees the part's pose relative to the fixture
and the true contact labels, and the contact supervision described next.
Randomizing dynamics and geometry \citep{Tobin2017DR,Peng2018Dynamics} covers
the ranges of \tabref{tab:ranges}.

\paragraph{Contact supervision.} A blind policy has to infer contact from how
the arm moves. To give that inference a dense training signal, we supervise
the actor's latent with the contact state it cannot observe. At every step,
the simulator labels points sampled on the mating features of the part and fixture, as free, touching but misaligned, or aligned
(\figref{fig:arch}, top; Appendix~\ref{app:training} details each task's labels).
A small contact decoder predicts these labels
from the latent $z_t$, and its class-weighted cross-entropy is added to the
PPO loss \citep{Jaderberg2017Auxiliary}. The gradient of this loss trains the
encoder, but the prediction never enters the policy, so nothing privileged
reaches the actor and the decoder is dropped at deployment.

\paragraph{Visual policies.} To isolate the sensing modality, we train
visual policies with the same task, resets, randomization, reward, actions,
and success check. A camera rigidly mounted on the wrist renders a
64$\times$64 RGB image of the held part and the fixture. The visual policy sees only
images, over the same 0.6\,s history, through a compact CNN.

\section{Results}
\label{sec:science}

We design experiments to answer the following questions:
\begin{enumerate}[leftmargin=3em]
    \item (\secref{sec:existence}) Can a policy that senses only its own joints learn precise assembly tasks including peg insertion, gear meshing, and nut threading?
    \item (\secref{sec:smart}) Do blind policies learn intelligent, contact-reactive search, or is it learning scripted patterns?
    \item (\secref{sec:unseen}) Do blind policies generalize better than visual policies to part geometries outside the training distribution?
    \item (\secref{sec:real}) Can blind policies that rely on contact-rich search be transferred zero shot with high fidelity to the real world?
\end{enumerate}

\subsection{Proprioception is sufficient}
\label{sec:existence}

\begin{figure}[t]
  \centering
  \includegraphics[width=\linewidth]{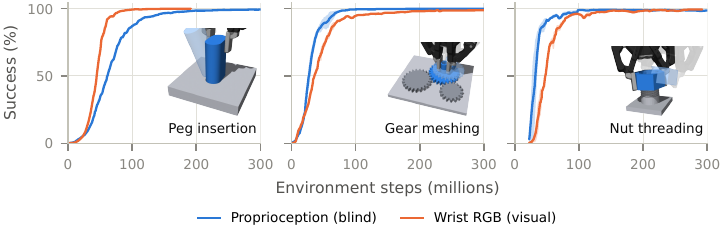}
  \vspace{-1.6em}
  \caption{\textbf{Proprioception alone is enough to learn all three tasks.} Blind
  policies, which sense only their own joints, reach the same near-perfect training
  success as visual policies with a wrist camera. Lines are the mean of two training seeds.}
  \label{fig:curves}
\end{figure}

All three tasks achieve near-perfect success on in-distribution geometries (\figref{fig:generalization}, left). The joint history alone therefore carries enough information about contact to find an
opening, bring teeth into mesh, and engage a thread.

\begin{figure}[t]
  \centering
  \includegraphics[width=\linewidth]{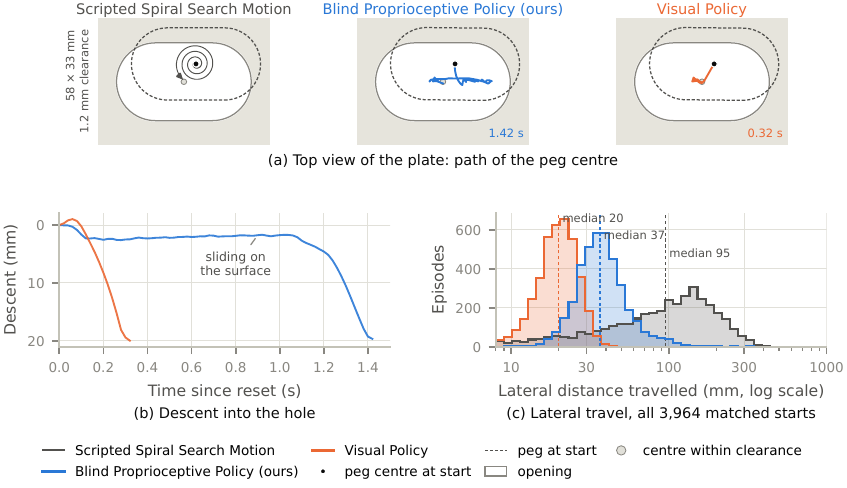}
  \vspace{-1.5em}
  \caption{\textbf{Proprioception guides efficient search without vision.} Because the blind policy must search through contact, it follows a less direct path and seats later than the visual policy (a, b). Nonetheless, it travels far less (c) than an optimistic scripted spiral search baseline
  \citep{Chhatpar2001Search}.}
  \label{fig:behavior}
\end{figure}

Compared with visual policies, which can typically observe the poses of the grasped and mating parts directly, blind policies may be slower. On peg insertion, the blind
policy has to search for the opening through contact, so
it takes a less direct path and seats later (\figref{fig:behavior}a, b) than a visual policy. The gap is
smaller on the other two tasks (Appendix~\ref{app:more}), plausibly because the grasped part occludes more of the mating part from the camera's view, so the visual policy, too, has to search to some extent.

\subsection{Visually blind but intelligent}
\label{sec:smart}

\begin{figure}[t]
  \centering
  \includegraphics[width=\linewidth]{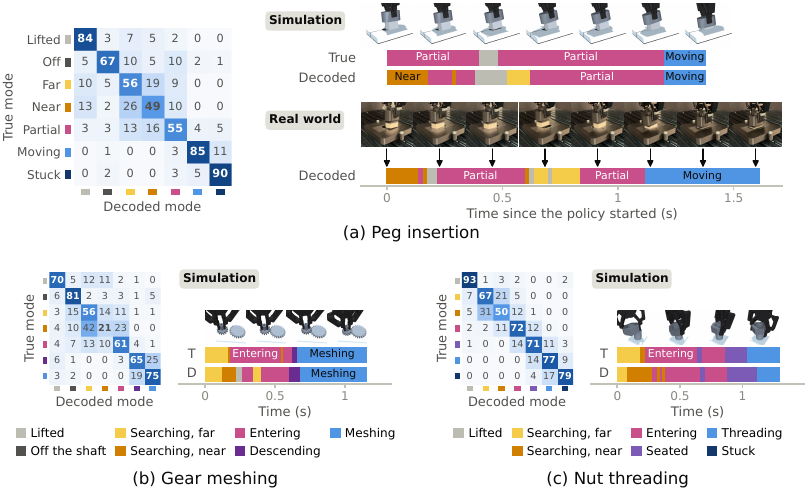}
  \vspace{-1.5em}
  \caption{\textbf{The blind policies' joint histories encode the task's mode, in
  simulation and in the real world.} A linear probe fitted only in simulation reads
  the mode from each policy's history latent. Each matrix gives its confusions (\%);
  beside it, one start state is decoded at every step (T: true mode, D: decoded). On
  all three tasks, the probe cleanly separates modes that call for different actions,
  such as lifted, engaged, and stuck. (a)~On the robot,
  it decodes the same sequence of peg insertion modes as in simulation.}
  \label{fig:probe}
\end{figure}

If blind policies search reactively rather than follow a script, their history
latent $z_t$, through which everything the actor knows about the past passes
(\figref{fig:arch}), must encode the state of contact. To test this, we roll out
each policy in simulation
and label every state with one of seven modes of its task, computed from the
privileged part pose. For the peg, these are \modebox{BDBCB2}lifted,
\modebox{52514E}beside the plate, searching \modebox{F5CB4A}far from the opening or
\modebox{D17F00}near it, \modebox{C94F8A}partly engaged, and fully inside, either
\modebox{4F95E3}moving or \modebox{13366B}stuck; the gear's and nut's modes follow
similar stages of their tasks (Appendix~\ref{app:probe}). We then fit a linear
probe that predicts the mode from $z_t$.

On all three tasks, the probe separates modes that call for different actions,
such as lifted, engaged, and stuck (\figref{fig:probe}), and does so more accurately
from $z_t$ than from the current observation alone (accuracy 61--73\%,
against 41--64\%; chance 14\%). Furthermore, the policy heavily relies on these inferred contact states to choose actions. Swapping in the latent of a
state in another mode changes the action 1.3--3.0$\times$ as much as a swap within
the same mode
(\figref{fig:probe-swap}; Appendix~\ref{app:probe}). Fitted only in simulation, the
peg probe decodes the same sequence of modes on the real robot (\figref{fig:probe}a).

Reacting to contacts also makes a blind assembly efficient. A fixed pattern
sweeps the same path regardless of what the arm feels, whereas a reactive policy can use the information gathered through contact, such as the peg catching the rim of the opening, to steer its
search. We test this on peg insertion, where a scripted blind baseline is standard:
spiral search, which moves the peg outward along a spiral until it drops into the
hole, is a common strategy for insertion under position uncertainty in industrial
assembly \citep{Chhatpar2001Search,Park2017Compliance,Marvel2018MultiRobot}. From
the same starts, the blind policy travels a median 37\,mm before seating, against 95\,mm for an optimistic
spiral that starts flush with the hole opening, and is given the peg's orientation and the best starting direction
(\figref{fig:behavior}c).

\subsection{Generalization to unseen parts}
\label{sec:unseen}

\begin{figure}[t]
  \centering
  \includegraphics[width=\linewidth]{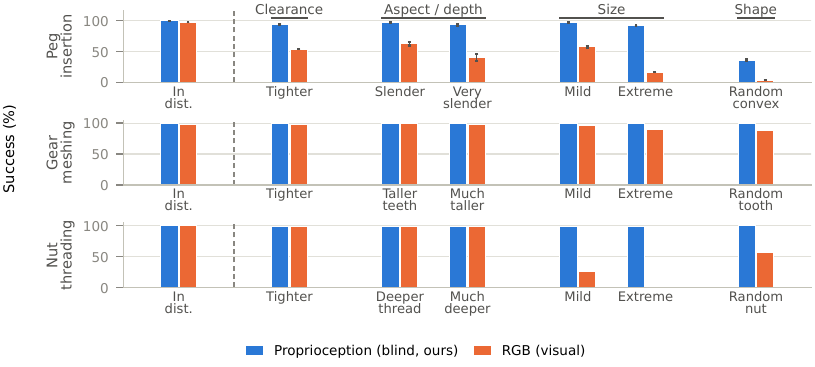}
  \vspace{-1.6em}
  \caption{\textbf{Proprioceptive policies remain more robust to geometric variation.} Success of the
  blind and visual policies (both with a 30-step history) on each task, in
  distribution (left of the dashed line) and on unseen parts, each widening one
  object property beyond its training range while the others stay within
  theirs.}
  \label{fig:generalization}
\end{figure}

On all three tasks, both blind policies and visual policies are
reliable on the parts they were trained on. However, the blind policy is more reliable on assemblies sampled from outside the training distribution  (\figref{fig:generalization}). The visual policy is particularly vulnerable to geometry changes that significantly change how the parts look; it consistently underperforms the blind policy across tasks when evaluated on parts that are of a different shape, or extremely different size, than the training distribution. The visual policy learns to align visual features which may not appear in held-out assemblies. The blind policy, on the other hand, learns a searching strategy that is much more transferable.

\subsection{Zero-shot transfer to the real world}
\label{sec:real}

\begin{figure}[t]
  \centering
  \includegraphics[width=\linewidth]{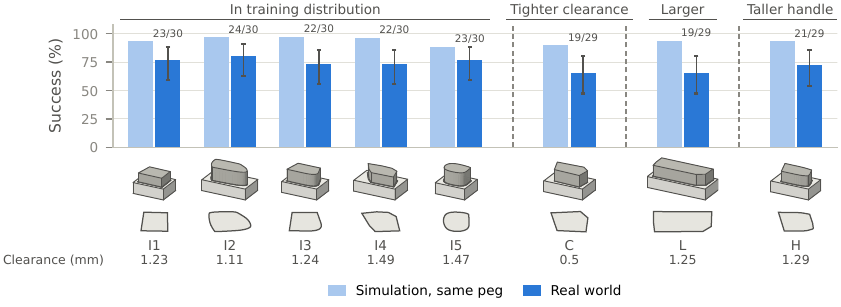}
  \vspace{-1.2em}
  \caption{\textbf{Real-world results} of the blind policy, with no
  fine-tuning, beside the same policy in simulation on the same peg (2,048
  episodes). Error bars are 95\% Wilson intervals. Details in
  Appendix~\ref{app:real}.}
  \label{fig:real}
\end{figure}

\paragraph{Setup.} We deploy the frozen blind policy on a Flexiv Rizon~4S with
a Robotiq 2F-85 at 50\,Hz, with no real-world data or fine-tuning. It receives
only encoder-derived joint and gripper state; no camera, force/torque sensor,
or hole pose enters the policy. The deployed policy is a policy
trained on random convex, smooth, and mixed cross-sections under a handle
(\tabref{tab:deployed}). We test eight 3D-printed pegs (\figref{fig:real}): five
in-distribution objects, and three that are drawn from outside the training distribution, a tighter clearance (0.5\,mm per side), a larger cross-section, and a
taller handle. Each trial starts from a
start state drawn from the simulated reset distribution, with the peg grasped
at a randomized position and tilt and touching the fixture near the hole. An operator marks a
trial as successful once the peg is seated.

\paragraph{Results.} One policy, with no parameter changes, succeeds in
173 of 237 trials (73\%; \figref{fig:real}), pressing onto
the surface, sweeping, and dropping in once the opening is under the peg, as in
simulation (\figref{fig:teaser}). It succeeds in 73--80\% of trials on each
in-distribution peg, and in 66\%, 66\%, and 72\% on the tighter, larger, and
taller-handled pegs, respectively.

\section{Related work}
\label{sec:related}

\paragraph{Robotic insertion.} Classical peg-in-hole assembly combines
compliance with search patterns such as spirals, rasters, and tilting
\citep{Whitney1982QuasiStatic,Chhatpar2001Search,Marvel2018MultiRobot}, in
some cases without force feedback \citep{Park2017Compliance}; these
strategies assume a known grasp and a hole within the pattern and are tuned
for each part. Learned insertion policies usually add force/torque sensing
\citep{Inoue2017HighPrecision,BeltranHernandez2020Variable}, wrist vision
\citep{Schoettler2020Industrial,Bogunowicz2020Eye,Spector2021InsertionNet,Haugaard2021Fast,Seo2025EquiContact},
or both \citep{Lee2020Making,Luo2021Robust,Watson2020Autonomous}. Wrist force/torque sensing combined with a soft wrist has allowed policies to tolerate 10\,mm of
hole-position error and an unseen peg shape \citep{Fuchioka2024SoftWrist},
while a blind dexterous policy relying on fingertip tactile sensing
\citep{Pan2026BeyondBinary} was demonstrated for precise insertion. Others
\citep{Tang2023IndustReal,Tang2024AutoMate,Noseworthy2024FORGE,Narang2022Factory} observe joint
state and a direct goal pose from perception or calibration with a few millimeters
of error. Our policy instead uses proprioception alone: it is given no
hole pose, part model, or force signal, and must locate the hole through
contact while the peg's pose in its grasp is unknown.

\paragraph{Proprioceptive policies.} Legged robots traverse rough terrain
with policies built on histories of onboard state
\citep{Hwangbo2019Agile,Lee2020Quadrupedal,Siekmann2021BlindStairs,Kumar2021RMA,Radosavovic2024Humanoid}.
In manipulation, proprioceptive policies reorient unseen objects in hand
\citep{Chen2021Reorientation,Qi2022InHand}, and \emph{Blind Dexterity} treats
joint and action histories as a whole-body tactile channel
\citep{Bhatt2026BlindDexterity}. These tasks involve forceful contact or
centimeter tolerances; insertion needs neither. In imitation learning, robot state can hurt
\citep{Ghosh2024Octo} by suppressing visual learning
\citep{Lu2026VisionProprio}.

\paragraph{Simulating tight contact.} Approximate convex decomposition
\citep{Mamou2009VHACD,Wei2022CoACD} tolerates concavity relative to the mesh
size, which can exceed an assembly's clearance. Exact polygon partitions are
classical \citep{Keil1985Decomposing}, and simulation benchmarks build holes
from primitives by hand
\citep{Zhu2020robosuite,Tao2024ManiSkill3,Zakka2025Playground}; \mi{}
automates this for arbitrary outlines and generates assemblies in bulk. SDFs
handle concavity directly and underlie GPU assembly simulation in Factory and
its successors \citep{Narang2022Factory,Macklin2020SDF,Tang2023IndustReal},
whose assets obtain their clearance approximately, by displacing mesh vertices
or eroding voxels \citep{Tang2024AutoMate,Wang2025MatchMaker}; SDF accuracy
depends on field resolution and contact search. Newton combines SDFs with
hydroelastic contact \citep{Newton}, and barrier methods simulate threads and
gear trains without interpenetration
\citep{Ferguson2021RigidIPC,Lan2022ABD}. Our benchmark (\secref{sec:sim})
follows the simulator comparisons of \citet{Erez2015Simulation} and
\citet{LeLidec2024Contact}. CRISP \citep{Lee2026CRISP} recently compared
engines on tight insertion and threading; we instead keep MuJoCo's solver
fixed and vary only the collision geometry, using passive probes with analytic
answers and speed--accuracy Pareto fronts.

\section{Limitations and conclusion}

We trained blind policies that insert pegs, mesh gears, and thread nuts from
proprioception alone, with no fixture pose, grasp, or part model, on thousands of
assemblies that \mi{} generates with exact collision geometry. The peg insertion
policy transfers zero-shot to the real world, and in simulation, blind policies
generalize to unseen parts far better than matched visual policies. The peg
insertion policy is, however, a last-mile manipulation skill, and its limits are
those of one. First, it assumes it starts in contact: every training episode
begins with the peg already touching the fixture a few centimeters from the
hole, so another motion, scripted or visual, must bring it there. Second, its
behavior is local. It searches the surface around where it starts and
corrects the misalignments it met in training, but it cannot handle events
that call for more than a local correction, such as a peg that slides out from the
grasp and lands far from the hole.

\subsubsection*{Acknowledgments}

This research was financially supported by the Ministry of Trade, Industry, and Energy
(MOTIE), Korea, under the ``Global Industrial Technology Cooperation Center program''
supervised by the Korea Institute for Advancement of Technology (KIAT). (Grant
No.~P0028435)

\bibliographystyle{iclr2026_conference}
\bibliography{references}

\newpage
\appendix

\section{Training details}
\label{app:training}

\paragraph{Study and deployed policies.} To cover as much real-world variation as
possible, the deployed peg policy (\secref{sec:real}) is trained on the full ranges of part geometry, physics parameters, and object pose that we can (limited by physics stability, geometric constraints like the gripper width, etc.). We also train similarly diverse
gear and nut policies for the latent probe (\secref{sec:smart};
Appendix~\ref{app:wide}). In order to test the policies' capacity for generalizing out of the training distribution, we design study training ranges for all three tasks (\secref{sec:science}) that are more narrow (\tabref{tab:ranges}), so that we can systematically widen each property beyond training
(\secref{sec:unseen}).

\subsection{Study policies}
\label{app:study}

\begin{table}[p]
  \centering
  \caption{\textbf{Study training ranges.} The study policies of all three tasks
  (\secref{sec:science}) train on these ranges; Appendix~\ref{app:wide} gives the
  wider ranges of the deployed peg policy and the full-range gear and nut policies.
  Ranges marked (log) are sampled log-uniformly and the rest uniformly. Geometry is
  drawn once per world; poses and contact parameters at every reset. No observation
  noise is added. $^\ddagger$The peg's hole center; the gear's and bolt's fixtures
  draw the same offsets from their own nominal positions.}
  \label{tab:ranges}
  \footnotesize
  \setlength{\tabcolsep}{3pt}
  \renewcommand{\arraystretch}{1.1}
  \begin{tabularx}{\linewidth}{@{}>{\raggedright\arraybackslash}p{3.3cm}YYY@{}}
    \toprule
    & Peg insertion & Gear meshing & Nut threading \\
    \midrule
    \multicolumn{4}{@{}l}{\emph{Part and fixture}} \\
    Part & stadium prism & involute spur gear, held by an 18\,mm boss; 1--2 anchor gears & ISO metric nut and bolt, coarse and fine pitch \\
    Size & scale 1.35--2.45; opening 38--91 $\times$ 20--54\,mm; length 50--60\,mm $\times$ scale & module 2.375--3.125\,mm; 16--24 teeth & M36 and M42 \\
    Shape & aspect ratio 1.25--2.5 (log), area-preserving & pressure angle 21.25--23.75$^\circ$; tooth width 0.45--0.47 pitch; face width 9.75--13.25\,mm & across flats and height 0.95--1.05 $\times$ ISO 4032 \\
    Clearance & 1.0--1.5\,mm per side & backlash 0.4--0.6\,mm per anchor; bore 0.75--1.0\,mm & flank 0.1--0.2\,mm (log) \\
    Fixture & hole 10\,mm $\times$ scale deep in an 80\,mm $\times$ scale square plate & shaft radius 4.5--5.5\,mm, 4.75--8.25\,mm above the gear & bolt stick-out 7.5--22.5\,mm above its minimum \\
    Part mass & 0.35--0.675\,kg & 128--202\,g with boss & 73--464\,g \\
    \midrule
    \multicolumn{4}{@{}l}{\emph{Contact}} \\
    Fixture friction (log) & plate 0.5--2.0 & plate and shaft 0.5--2.0; anchor teeth 0.30--0.67 & thread 0.24--0.34 \\
    Fixture contact time constant (log) & 3.5--7.1\,ms & 3.5--7.1\,ms & thread 2\,ms, fixed \\
    Anchor hinge (log) & --- & damping 0.32--3.2\,mN\,m\,s; friction loss 0.27--1.9\,mN\,m & --- \\
    Finger friction scale (log) & \multicolumn{3}{c}{1.41--2.0} \\
    Finger contact time-constant scale (log) & \multicolumn{3}{c}{0.78--1.34} \\
    Gripper force cap scale (log) & \multicolumn{3}{c}{1.41--2.83} \\
    \midrule
    \multicolumn{4}{@{}l}{\emph{Poses}} \\
    Fixture center from robot base & \multicolumn{3}{c}{$x$ 0.54--0.76\,m, $y$ $-0.29$--0.17\,m, $z$ $-0.20$--0.29\,m$^\ddagger$} \\
    Fixture tilt & roll, pitch, yaw $\pm5^\circ$ & roll, pitch, yaw $\pm5^\circ$ & roll, pitch $\pm5^\circ$ \\
    Gripper rotation, per axis & \multicolumn{3}{c}{$\pm5^\circ$} \\
    Initial part position, per axis & $\pm12.5$\,mm from the hole center & $\pm7.5$\,mm from the shaft axis & $\pm0.42$ bolt diameters from its axis \\
    Initial phase & --- & anchors at random angles & 87.5--142.5$^\circ$ of turning before the thread catches \\
    Grasp position & $\pm5.5$\,mm along the pads & $\pm2.5$\,mm along the pads; pad bottoms 6.75--10.25\,mm above the gear & $\pm1$\,mm across the pads; middle half of the pads' height \\
    In-hand tilt about the finger axis & $\pm10^\circ$ & $\pm5^\circ$ & $\pm10^\circ$ \\
    \midrule
    \multicolumn{4}{@{}l}{\emph{Control}} \\
    Joint gains $k_p$ / $k_d$; action scale & 80 / 20; 0.02\,rad & 40 / 20; 0.04\,rad & 40 / 20; 0.04\,rad (wrist joint: 0.2\,rad) \\
    Joint speed and torque penalty & none & none & none \\
    \bottomrule
  \end{tabularx}
\end{table}

\paragraph{Tasks.}
\label{app:gear-nut}
Physics runs at 1\,kHz under a 50\,Hz policy, and each world holds one part and
fixture drawn from \mi{}'s generators. \emph{Peg insertion:} each episode starts
with the peg touching the fixture's top face, a median 1\,cm from the hole center.
It succeeds when the entire peg bottom face is seated in the hole. Bonuses of $+1$
reward the peg's first capture by the opening and success. Contact labels mark each
of 64 points on the peg's bottom edge as free (more than 1\,mm above the fixture's
top face), in contact (on the rim or top face, within 1\,mm), or inside the opening.
An episode lasts at most 5\,s. \emph{Gear meshing:} each world holds a spur-gear set
from \mi{}'s gear generator, and the robot holds the gear by a printed boss. Each
episode starts with the gear resting on the shaft tip, off the shaft axis, with its
teeth out of phase with the anchors'. It succeeds when the bore surrounds the shaft
and the gear sits within 1\,mm of the shaft's collar, which the anchors' teeth allow
only in mesh. Bonuses reward the first capture of the shaft, the first mesh with
every anchor, and success; contact labels mark 64 points on the shaft's rim (free,
on the gear, or inside the bore) and 64 on the gear's tooth tips (free, on an
anchor, or in mesh); an episode lasts at most 10\,s. \emph{Nut threading:} each
world holds an ISO metric nut and bolt pair from \mi{}'s thread generator. Each
episode starts with the nut resting on the bolt tip, tilted and off the bolt axis,
short of the turn at which the thread catches it. It succeeds a quarter turn after
the thread catches, with the nut within 1\,mm of the bolt axis; bonuses reward
seating, capture, each new extent of the turn, and success; contact labels mark 64
points on the rim of the nut's bore (free, on the bolt's tip, around the bolt, or
caught by the thread, within a band of 0.29 pitch, 1\,mm at M30$\times$3.5); an
episode lasts at most 15\,s. Every episode also ends when the part moves far from
the fixture, and a nut episode when the nut leaves the grasp.

\paragraph{Architecture.} The actor observes 33 values per step: seven joint
positions, seven joint velocities, the gripper pose and twist (twelve), and its
previous action (seven). The nut policy's actor encodes the gripper orientation by
two columns of its rotation matrix rather than by Euler angles (36 values), since
the wrist turns through half a revolution while threading. A temporal convolutional
encoder maps the last 30 observations (0.6\,s) to a 32-dimensional latent; an MLP
(256, 256, 128 units) maps the latent and the current observation to seven
joint-position increments, scaled by the action scale of \tabref{tab:ranges} and
added to the measured joint positions. The critic sees the part's pose relative to
the fixture, the contact labels, and, for the nut, the thread's geometry (240, 434,
and 317 inputs for the peg, gear, and nut).

\paragraph{PPO.} 12,288 worlds, 32-step rollouts, 5 epochs of 20 minibatches (24
for the gear and nut), fixed learning rate $3\times10^{-4}$, clip 0.2,
$\gamma=0.98$, $\lambda=0.95$, entropy coefficient 0.003, value-loss coefficient 1
with clipping, gradient norm clipped at 1, and initial action standard deviation
0.5. The contact decoder has two hidden layers of 128 units and class weights 0.25
(free), 3.0 (contact), and 2.0 (inside, in mesh, around the bolt, or caught). The
visual policies see 64$\times$64 wrist RGB images over the same 30 steps.

\subsection{Deployed and full-range policies}
\label{app:wide}

\paragraph{Deployed peg policy.} The policy we deploy (\secref{sec:real}) trains as
the study peg policies do, on the ranges of \tabref{tab:deployed}: random convex
cross-sections under a handle, and wider contact and pose ranges. Its
starts lie a median 2\,cm from the hole center. It uses the gear's and nut's
controller and a staged safety penalty, which starts once the task has been learned
without safety limits and penalizes joint speeds and commanded torques above one
sixth of the hardware limits. PPO uses 24 minibatches and an exploration anneal,
during which the learning rate adapts to a target KL of 0.01.

\begin{table}[!htbp]
  \centering
  \caption{\textbf{Deployed peg policy's training ranges}, where they differ from the
  study peg's (\tabref{tab:ranges}); scale, aspect ratio, and clearance are the
  study's. $^\dagger$Polygonal, smooth, and mixed outlines in equal shares, each
  inside a stadium of the given scale and aspect ratio and redrawn until its hole
  fits the printed fixture insert (at most 106$\times$52\,mm), under an
  80$\times$44$\times$40\,mm stadium handle.}
  \label{tab:deployed}
  \small
  \setlength{\tabcolsep}{4pt}
  \begin{tabular}{ll}
    \toprule
    Parameter & Deployed peg policy \\
    \midrule
    \multicolumn{2}{l}{\emph{Part and fixture}} \\
    Cross-section & random convex$^\dagger$ \\
    Opening, long $\times$ short side & 39--91 $\times$ 30--52\,mm \\
    Peg length & 25--35\,mm, below the handle \\
    Hole depth & 20\,mm \\
    Plate & convex, 5--70\,mm margin \\
    Peg mass & 0.12--0.29\,kg with handle \\
    \midrule
    \multicolumn{2}{l}{\emph{Contact}} \\
    Plate sliding friction (log) & 0.25--4.0 \\
    Plate contact time constant (log) & 2.5--10\,ms \\
    Finger friction scale (log) & 1.0--2.0 \\
    Finger contact time-constant scale (log) & 0.6--1.75 \\
    Gripper force cap scale (log) & 1.0--4.0 \\
    \midrule
    \multicolumn{2}{l}{\emph{Poses}} \\
    Fixture center from robot base, $x$ & 0.39--0.83\,m \\
    \hspace{1em}$y$ / $z$ & $-0.46$--0.46 / $-0.48$--0.43\,m \\
    Fixture roll, pitch, yaw & $\pm10^\circ$ \\
    Gripper rotation, per axis & $\pm10^\circ$ \\
    Initial peg position, per axis & $\pm25$\,mm \\
    Grasp position along the pads & $\pm7$\,mm, on the handle \\
    In-hand tilt about the finger axis & $\pm20^\circ$ \\
    \midrule
    \multicolumn{2}{l}{\emph{Control}} \\
    Joint gains $k_p$ / $k_d$; action scale & 40 / 20; 0.04\,rad \\
    Joint speed and torque penalty & staged \\
    \bottomrule
  \end{tabular}
\end{table}

\paragraph{Full-range gear and nut policies.}
\label{app:full-range}
The gear and nut probes of \secref{sec:smart} use policies trained on the full
ranges. Gears have a module of 2--3.5\,mm and 12--28 teeth,
pressure angles of 20--25$^\circ$, tooth widths of 0.44--0.48 of the pitch, face
widths of 8--15\,mm, 0.2--0.6\,mm of backlash per anchor, and 0.5--1\,mm of bore
clearance, on 4--6\,mm shafts 3--10\,mm above the gear; with the boss they weigh
128--276\,g, set by their geometry. Nuts are 870 ISO metric nut and bolt pairs,
M24--M42 at coarse and fine pitch, with 0.05--0.2\,mm of flank clearance, across
flats and heights within 10\% of ISO 4032, and steel masses of 73--855\,g. Their
contact and pose ranges are wider too: anchor-tooth friction 0.2--1.0, thread
friction 0.2--0.4, and the deployed peg policy's plate, finger, gripper, fixture,
and gripper-rotation ranges. Gears start a median 12\,mm off the shaft axis; nuts
start a median 19\,mm off the bolt axis and 60--170$^\circ$ of turning short of the
thread catching. With deterministic actions, the gear policy succeeds in 98\% of
8,192 episodes on 4,096 unseen gear sets drawn from its training distribution, and
the nut policy in 93\% of 8,192 episodes on 512 unseen nut and bolt pairs drawn
likewise.

\section{Collision benchmark details}
\label{app:bench}

\begin{table}[h]
  \centering
  \caption{\textbf{Passive probes.} Errors are in micrometers of wall position;
  negative values mean that phantom material stopped the peg early.}
  \small
  \begin{tabular}{lp{5.2cm}p{4.3cm}}
    \toprule
    Probe & Setup & Exact answer \\
    \midrule
    Wall offset & Rays from the hole axis at 720 angles and 11 heights; no solver & Wall on the analytic outline \\
    S0 contact onset & Aligned peg placed at known gaps; collision only & Contact appears at gap 0 \\
    S1 drop & centered free peg dropped from 1\,mm; $c$ from $+1$\,mm to $-500$\,\textmu m & Passes with zero wall contact iff $c>0$ \\
    S2 lateral play & Planar slides; 0.1\,N push in 24 directions & Support of the feasible translation set \\
    S3 tilt play & Slides and one hinge; 1.5\,mN\,m torque & Maximum tilt at which both faces fit \\
    S4 yaw play & Slides and a vertical hinge (square hole) & $\arcsin(a/(s\sqrt2))-\pi/4$ \\
    \bottomrule
  \end{tabular}
\end{table}

\begin{table}[h]
  \centering
  \caption{\textbf{Real-time factor} at a 1\,ms time step with the peg pressed
  against the wall; ranges span each method's settings, best of multi-point
  collision on and off. GPU figures sum over 512 MuJoCo Warp worlds on one
  RTX 5090. The last column gives the worst settled error at a 20\,\textmu m
  radial clearance.}
  \label{tab:speed}
  \small
  \setlength{\tabcolsep}{4pt}
  \begin{tabular}{lrrl}
    \toprule
    Representation & CPU, 1 thread & GPU, 512 worlds & Error at $c=20$\,\textmu m \\
    \midrule
    \mi{}, square hole & 93$\times$ & 970$\times$ & 0.2\,\textmu m \\
    \mi{}, circle, 7--32 sides & 82--29$\times$ & 1120--1070$\times$ & $\le$2.6\,\textmu m \\
    \mi{}, circle, 63--281 sides & 16--4.9$\times$ & 875--442$\times$ & $\le$0.4\,\textmu m \\
    Mesh SDF, octree depth 5--8 & 2.9--2.2$\times$ & 590--840$\times$ & fills, or 23--79\,\textmu m \\
    CoACD, threshold 0.01 (839 hulls) & 1.8$\times$ & 254$\times$ & fills \\
    Other CoACD and V-HACD settings & 3.3--70$\times$ & 680--1900$\times$ & fills \\
    \bottomrule
  \end{tabular}
\end{table}

All scenes use a 1\,ms time step, the implicit-fast integrator, the Newton
solver, $\mathrm{solref}=(2\,\mathrm{ms},1)$, $\mathrm{solimp}=(0.99,0.999,0.001)$,
$\mu=1$, and zero margin. Baselines receive an exact, watertight plate mesh
(the circle tessellated at 1,024 segments, 0.04\,\textmu m sagitta), and ground
truth uses the analytic circle, so \mi{} is charged for its own chord error.
The peg is always an exact MuJoCo primitive, so every difference comes from
the hole. Timings come from one idle machine (Intel i9-13900KS, NVIDIA RTX
5090). Warp contact and constraint buffers are calibrated per representation,
and every GPU run is checked for buffer overflow, which would silently drop
contacts. At a 2\,ms contact time constant, frictionless drops let pegs up to
20\,\textmu m oversize slip in for every method, including exact geometry, so
S1 cannot resolve geometric errors below about 20\,\textmu m; the play probes
can.

\paragraph{Setup.} The round and square holes are 16\,mm wide in a 15\,mm
plate, with radial clearance $c$ from 20 to 200\,\textmu m; at
$c=20$\,\textmu m, $2c/D=0.0025$. The M10$\times$1.5 bolt and nut have flank
clearance $c$ of 20 and 100\,\textmu m and are probed for axial, radial, and
tilt play and for screwing under load. CPU times use one environment on one
thread; GPU times use 512 MuJoCo Warp worlds, except in
\figref{fig:pareto}b, which uses 256 and measures the error on Warp as well.

\section{Additional results}
\label{app:more}

\begin{figure}[h]
  \centering
  \includegraphics[width=\linewidth]{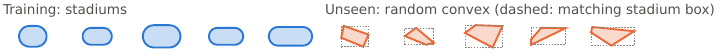}
  \vspace{-1.4em}
  \caption{\textbf{Study peg cross-sections, to scale.} The study policies train
  on stadiums; the unseen random convex pegs used only for evaluation have one
  vertex on each side of a matching stadium's bounding box and the same
  clearance.}
  \label{fig:shapes}
\end{figure}

\begin{table}[h]
  \centering
  \caption{\textbf{In-distribution behaviour} (\twosecrefs{sec:existence}{sec:smart}), from
  8,192 episodes per seed, averaged over both seeds (success, time, rate), and from
  the 3,964 of 4,096 matched starts where both seed-1 policies succeed (travel,
  engagement, tilt). Medians
  unless stated; insertions per minute count every attempt, with failures at
  their full 5\,s. Starts lie a median 10\,mm from the hole center.}
  \label{tab:behavior}
  \small
  \begin{tabular}{lcc}
    \toprule
    & Blind proprioceptive (ours) & Visual \\
    \midrule
    Success (\%) & 99.4 & 97.2 \\
    Time to seat (s) & 0.80 & 0.44 \\
    Insertions per minute & 65 & 82 \\
    Lateral travel before seating (mm) & 37 & 20 \\
    Time partly engaged (s) & 0.38 & 0.22 \\
    Tilt when partly engaged / seated ($^\circ$) & 6.7 / 1.1 & 5.2 / 1.1 \\
    \bottomrule
  \end{tabular}
\end{table}

\paragraph{Behaviour.} The blind policy travels farther than the visual
policy from 99\% of matched starts (\tabref{tab:behavior}). On unseen random
convex pegs, the blind policy succeeds in 36.2\% of episodes and the visual
policy in 3.9\% (both seeds); these pegs have a median 53\% of the matching stadium's area.

\paragraph{Speed on gear meshing and nut threading.} On the in-distribution episodes of
the study policies of \figref{fig:generalization} (8,192 per policy, one seed,
deterministic actions), counting successful episodes only, the blind gear policy
succeeds in a median 0.58\,s against the visual policy's 0.64\,s, and takes longer on
43\% of the 8,059 starts from which both succeed. Plausibly, the held gear hides the
shaft from the wrist camera while it aligns, so visual policies also have to search. The blind nut
policy succeeds in a median 0.76\,s against 0.64\,s, and takes longer on 98\% of the
8,139 starts from which both succeed (by a median 0.10\,s).

\begin{figure}[t]
  \centering
  \includegraphics[width=\linewidth]{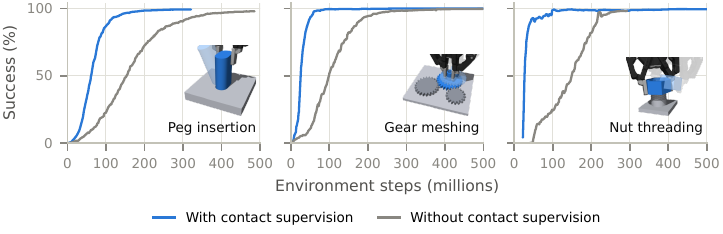}
  \vspace{-1.6em}
  \caption{\textbf{Contact supervision speeds up learning.} Training success of the
  blind study policies of \figref{fig:curves} (blue) and of the same policies trained
  without the contact loss (gray).}
  \label{fig:aux}
\end{figure}

\paragraph{Contact supervision.}
\label{app:aux}
Without the contact loss (\secref{sec:policy}), the blind policies still learn all
three tasks, but need 2.7 (peg), 3.8 (gear), and 5.1 (nut) times as many environment
steps to reach 90\% training success (\figref{fig:aux}). The unsupervised peg policy
reaches 97.8\% after 486M steps; the supervised policies pass 99\% by 242M.

\begin{figure}[t]
  \centering
    \includegraphics[width=\linewidth]{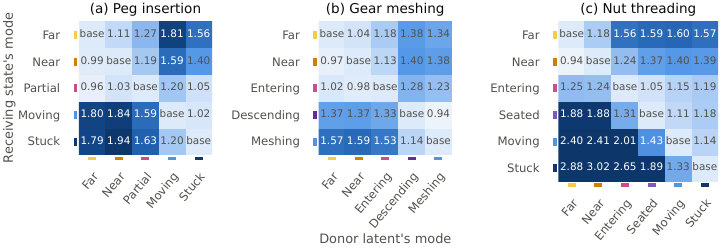}
  \vspace{-0.6em}
  \caption{\textbf{Latent-swap test} for the policies of \figref{fig:probe} in simulation.
  Each state keeps its own current observation but receives the history
  latent $z_t$ of a state in another mode; each cell is the mean change in
  action relative to a swap within the receiving state's own mode.
  }
  \label{fig:probe-swap}
\end{figure}

\paragraph{Probe details.}
\label{app:probe}
The search modes are split at 5\,mm from the opening, and an inside peg is
stuck when it progresses less than 0.5\,mm in 0.2\,s. The probe uses the
deployed policy (\secref{sec:real}). We roll it out in 1,536 worlds of its training
distribution (1.15M states), train the probes on even-numbered worlds, and
test on odd ones. Balanced accuracy over the seven modes is 70\% (chance 14\%)
from the history latent, against 56\% from the current observation alone and
33\% from a randomly initialized encoder. The history latent recognizes
lifted, moving, and stuck pegs with 84--90\% recall; it decodes far search as
near 19\% of the time, and near as far 26\%. Swapping latents between a
searching and an inside state changes the action 1.4--1.9$\times$ as much as
swapping between two states of the same mode; swapping between near and far
changes it 0.99--1.11$\times$ as much (\figref{fig:probe-swap}a). On the
robot, we run the policy in 30 further trials on peg I3 (\figref{fig:real}); its actions
recomputed from the logged observations match the logged ones to within
$2\times10^{-5}$. At the end of successful trials, the peg pose estimated from
the joint encoders and planned grasp is a median 6.6\,mm (10th--90th
percentile 2.6--17.4\,mm) from the seated pose, as the peg slips and turns in
the fingers, and places only 3\% of their states fully inside the hole. Scored
states have every point of the peg's bottom edge at least 3\,mm above the
plate (lifted) or at least 3\,mm inside the opening (inside). The estimate
never places the peg clearly above the plate; the probe labels 62\% of the
inside states correctly, against 100\% in simulation from the same start
states, and labels the final state of 17 of the 24 successful trials as
inside. In \figref{fig:probe}a, true and decoded modes of the start
state shown agree in simulation on 74\% of steps (median over the 30 start states:
65\%).

The gear and nut probes use the full-range policies of Appendix~\ref{app:full-range}
(PPO updates 3,085 and 2,000) with the same recipe: one episode with deterministic
actions in each of 1,536 worlds of held-out geometries from their training
distributions (133k and 85k states), probes trained on even-numbered worlds and
tested on odd ones. The modes come from the environments' contact labels and the
part's pose: lifted; off the shaft (gear only; the nut leaves its bolt too rarely to
probe); resting on the shaft or bolt tip and searching, far or near (split at the
median start distance, 7.2 and 19.7\,mm); entering, with the tip inside the bore but
not yet captured; for the gear, descending the shaft before its teeth reach the
anchors, then meshing, with its teeth on the anchors' tops or in mesh; for the nut,
seated on the bolt before the thread catches, then threading, or stuck when it
advances less than 0.05 pitch in 0.2\,s. Balanced accuracy over the seven modes is
61\% for the gear and 73\% for the nut (chance 14\%), against 41\% and 64\% from
the current observation and 29\% and 46\% from a randomly initialized encoder;
near search is decoded as far 42\% and 31\% of the time. Swapping latents between a
searching and an engaged state changes the action 1.3--1.6$\times$ (gear) and
1.4--3.0$\times$ (nut) as much as swapping between two states of the same mode;
swapping between near and far changes it 0.97--1.04$\times$ and 0.94--1.18$\times$
as much (\figref{fig:probe-swap}b, c). The start states shown come from a re-run of each rollout, decoded with
the same probes and rendered from the same states. Among the episodes that end
engaged and visit at least five modes, the nut's agrees with the true modes on
74\% of steps (the median), and the gear's on 79\% (the 75th percentile; median
69\%).

\paragraph{Unseen parts.}
\label{app:unseen}
Each unseen part of \figref{fig:generalization} widens one property of the part beyond
its study training range (\tabref{tab:ranges}) while every other range stays at its
study value. \tabref{tab:unseen} lists the widened ranges for each task, grouped as in
the figure: a tighter fit; a mild and a strong change of the part's proportions; a mild
and an extreme change of size, on both sides of the training range; and a new shape.
\tabref{tab:generalization} lists every bar.

\begin{table}[h]
  \centering
  \caption{\textbf{Unseen parts} of \figref{fig:generalization}. Each row widens one
  property beyond its study training range (first row of each group); the bar's name in
  the figure is in italics where it differs by task. Size bins pool parts smaller and
  larger than training.}
  \label{tab:unseen}
  \footnotesize
  \setlength{\tabcolsep}{3pt}
  \renewcommand{\arraystretch}{1.1}
  \begin{tabularx}{\linewidth}{@{}>{\raggedright\arraybackslash}p{1.6cm}YYY@{}}
    \toprule
    & Peg insertion & Gear meshing & Nut threading \\
    \midrule
    \multicolumn{4}{@{}l}{\emph{Clearance}} \\
    Training & 1.0--1.5\,mm per side & backlash 0.4--0.6\,mm per anchor & flank 0.1--0.2\,mm \\
    Tighter & 0.5--0.625\,mm & 0.2--0.25\,mm & 0.05--0.0625\,mm \\
    \midrule
    \multicolumn{4}{@{}l}{\emph{Aspect / depth}} \\
    Training & aspect ratio 1.25--2.5 & tooth height 1.0 module above the pitch circle & ISO thread depth \\
    Mild & \emph{Slender:} 3.54--4.20 & \emph{Taller teeth:} 1.1--1.2 modules & \emph{Deeper thread:} 1.1--1.2$\times$, same flank angle \\
    Strong & \emph{Very slender:} 4.20--5.0 & \emph{Much taller:} 1.2--1.3 modules & \emph{Much deeper:} 1.2--1.3$\times$ \\
    \midrule
    \multicolumn{4}{@{}l}{\emph{Size}} \\
    Training & scale 1.35--2.45 & module 2.375--3.125\,mm & M36 and M42 \\
    Mild & 1.075--1.35 or 2.45--2.725 & 2.19--2.375 or 3.125--3.31\,mm & M30 or M48 \\
    Extreme & 0.8--1.075 or 2.725--3.0 & 2.0--2.19 or 3.31--3.5\,mm & M24 or M52, the largest the gripper holds \\
    \midrule
    \multicolumn{4}{@{}l}{\emph{Shape}} \\
    Training & stadium & involute teeth & hexagon \\
    Random & \emph{Random convex:} one vertex on each side of a matching stadium's bounding box, same clearance (\figref{fig:shapes}) & \emph{Random tooth:} each tooth a random quadrilateral inside the involute one & \emph{Random nut:} a convex outline of 5--8 sides with two parallel gripped faces, 0.75--1.35$\times$ the hexagon's area \\
    \bottomrule
  \end{tabularx}
\end{table}

\begin{table}[h]
  \centering
  \caption{\textbf{Success on unseen parts} (\%), every bar of
  \figref{fig:generalization}; \tabref{tab:unseen} names each task's columns. Peg
  insertion is the mean over two training seeds per policy; gear meshing and nut
  threading are one seed. Both policies have a 30-step history. Best per task and column
  in bold.}
  \label{tab:generalization}
  \small
  \setlength{\tabcolsep}{3.5pt}
  \begin{tabular}{lccccccc}
\toprule
 & & Clearance & \multicolumn{2}{c}{Aspect / depth} & \multicolumn{2}{c}{Size} & Shape \\
\cmidrule(lr){3-3}\cmidrule(lr){4-5}\cmidrule(lr){6-7}\cmidrule(l){8-8}
Senses & In dist. & Tighter & Mild & Strong & Mild & Extreme & Random \\
\midrule
\multicolumn{8}{l}{\emph{Peg insertion}} \\
Proprioception (blind) & \textbf{99.4} & \textbf{93.8} & \textbf{97.1} & \textbf{93.3} & \textbf{97.1} & \textbf{91.9} & \textbf{36.2} \\
RGB (visual) & 97.2 & 54.1 & 62.7 & 40.1 & 57.7 & 16.4 & 3.9 \\
\midrule
\multicolumn{8}{l}{\emph{Gear meshing}} \\
Proprioception (blind) & \textbf{$>$99.9} & \textbf{$>$99.9} & \textbf{$>$99.9} & \textbf{$>$99.9} & \textbf{$>$99.9} & \textbf{$>$99.9} & \textbf{$>$99.9} \\
RGB (visual) & 98.4 & 98.0 & 98.6 & 98.5 & 96.5 & 90.2 & 87.9 \\
\midrule
\multicolumn{8}{l}{\emph{Nut threading}} \\
Proprioception (blind) & \textbf{99.7} & \textbf{99.2} & \textbf{99.2} & 98.2 & \textbf{98.8} & \textbf{98.4} & \textbf{99.7} \\
RGB (visual) & \textbf{99.7} & 98.7 & 99.1 & \textbf{98.8} & 25.5 & 1.6 & 57.1 \\
\bottomrule
\end{tabular}

\end{table}

\paragraph{Real-world pegs.}
\label{app:real}
The pegs of \figref{fig:real} are 3D printed under an 80$\times$44\,mm stadium
handle 40\,mm tall, except H, whose handle is 70\,mm tall; they weigh
147--164\,g, and every hole is 20\,mm deep. I1--I5 are cross-sections drawn
from the deployed policy's training distribution (polygonal, smooth, and mixed
outlines) with 1.11--1.49\,mm of clearance per side. C is a polygonal peg with
0.5\,mm of clearance; L enlarges a training outline by 1.28$\times$1.09 to a
101$\times$37\,mm hole; H puts a mixed outline under the taller handle. Each
peg has 30 planned resets, and each reset counts once, by its first attempt
that reached a verdict: the operator's success or failure, or the 20\,s
horizon, which counts as a failure. Each simulated rate is the same policy over 2,048 worlds of that
peg's geometry at the same 20\,s horizon, with deterministic actions.

\end{document}